\documentclass{article}

\usepackage{amsmath,amssymb,amsthm}
\usepackage{graphicx}
\usepackage{algorithm}
\usepackage{algorithmic}
\usepackage{booktabs}
\usepackage{longtable}
\usepackage{multirow}
\usepackage{soul}
\usepackage{hyperref}
\usepackage{natbib}
\usepackage{xcolor}
\usepackage{tikz}
\usepackage{subfigure}

\usepackage[preprint]{neurips_2025}

\usepackage[utf8]{inputenc} 
\usepackage[T1]{fontenc}    
\usepackage{hyperref}       
\usepackage{url}            
\usepackage{booktabs}       
\usepackage{amsfonts}       
\usepackage{nicefrac}       
\usepackage{microtype}      
\usepackage{xcolor}         

\title{PAME-AI: Patient Messaging Creation and Optimization using Agentic AI}

\author{%
  Junjie Luo \\
  Johns Hopkins School of Medicine \\
  Baltimore, MD, USA \\
  \texttt{jluo41@edu.edu} \\
  \And
  Yihong Guo \\
  Johns Hopkins University \\
  Baltimore, MD, USA  \\
  \texttt{yguo80@jhu.edu} \\
  \And
  Anqi Liu \\
  Johns Hopkins University \\
  Baltimore, MD, USA  \\
  \texttt{aliu74@jhu.edu} \\
  \AND
  Ritu Agarwal \\
  Johns Hopkins University, CDHAI \\
  Baltimore, MD, USA  \\
  \texttt{ritu.agarwal@jhu.edu} \\
  \And
  Gordon Gao \\
  Johns Hopkins University, CDHAI\\
  Baltimore, MD, USA  \\
  \texttt{gordon.gao@jhu.edu} \\\
}

\begin{document}

\maketitle

\begin{abstract}
Messaging patients is a critical part of healthcare communication, helping to improve things like medication adherence and healthy behaviors. However, traditional mobile message design has significant limitations due to its inability to explore the high-dimensional design space. We develop PAME-AI, a novel approach for \textbf{Pa}tient \textbf{Me}ssaging Creation and Optimization using Agentic \textbf{AI}. Built on the Data-Information-Knowledge-Wisdom (DIKW) hierarchy, PAME-AI offers a structured framework to move from raw data to actionable insights for high-performance messaging design. PAME-AI is composed of a system of specialized computational agents that progressively transform raw experimental data into actionable message design strategies. 
We demonstrate our approach's effectiveness through a two-stage experiment, comprising of 444,691 patient encounters in Stage 1 and 74,908 in Stage 2. The best-performing generated message achieved 68.76\% engagement compared to the 61.27\% baseline, representing a 12.2\% relative improvement in click-through rates. This agentic architecture enables parallel processing, hypothesis validation, and continuous learning, making it particularly suitable for large-scale healthcare communication optimization. 

\end{abstract}

\section{Introduction}

With increasing digitalization of healthcare services, the effectiveness of patient engagement through mobile messaging directly impacts health outcomes, medication adherence, and healthcare system efficiency \cite{milkman2022680, thakkar2016mobile, janz1984health}. 
However, optimizing such communication requires processing vast amounts of heterogeneous data, understanding complex behavioral patterns, and generating actionable insights that can be implemented at scale. Message design often faces the curse of dimensionality. Consider the sheer number of message variations. 
With just ten attributes (such as information richness, portion of clinical knowledge, framing, urgency, etc), each having three levels (e.g., urgency: low, medium, high), there are $3^{10}$
 =59,049  possible versions. 
As the possible versions of messages grow exponentially with dimensions, exhaustively testing each one to find the most effective message is often not feasible. 
Optimizing healthcare communication is thus a critical challenge at the intersection of behavioral science, data analytics, and system design.  

Traditional approaches to message optimization typically rely on manual hypothesis generation based on psychology, followed by A/B testing  \cite{milkman2024megastudy} or single machine learning models. This approach is severely constrained by the number of messages it can test, typically limited to single digit versions of messages. However, with such small number of candidates, the traditional approach struggles to uncover the most effective message. \cite{chiam2024nudgerank}. Also, there are no mechanism for future experiments to learn from previous ones. 

In this paper, we propose \textbf{PAME-AI}, a novel approach for \textbf{Pa}tient \textbf{Me}ssaging Creation and Optimization using Agentic \textbf{AI}. Our approach addresses the above  challenges by structuring the optimization process according to the Data-Information-Knowledge-Wisdom (DIKW) hierarchy \cite{ackoff1989data}. By doing so, PAME-AI provides the following capabilities: (1) systematically organize insights at different levels of abstraction (including learning from prior experiments), 
(2) improve the transparency of message generation process, and (3) balance exploration of new strategies with exploitation of proven approaches. PAME-AI can automatically generate large number of high-performance messages that outperform manual human experts.

\begin{figure}[htbp]
\centering
\includegraphics[width=\textwidth]{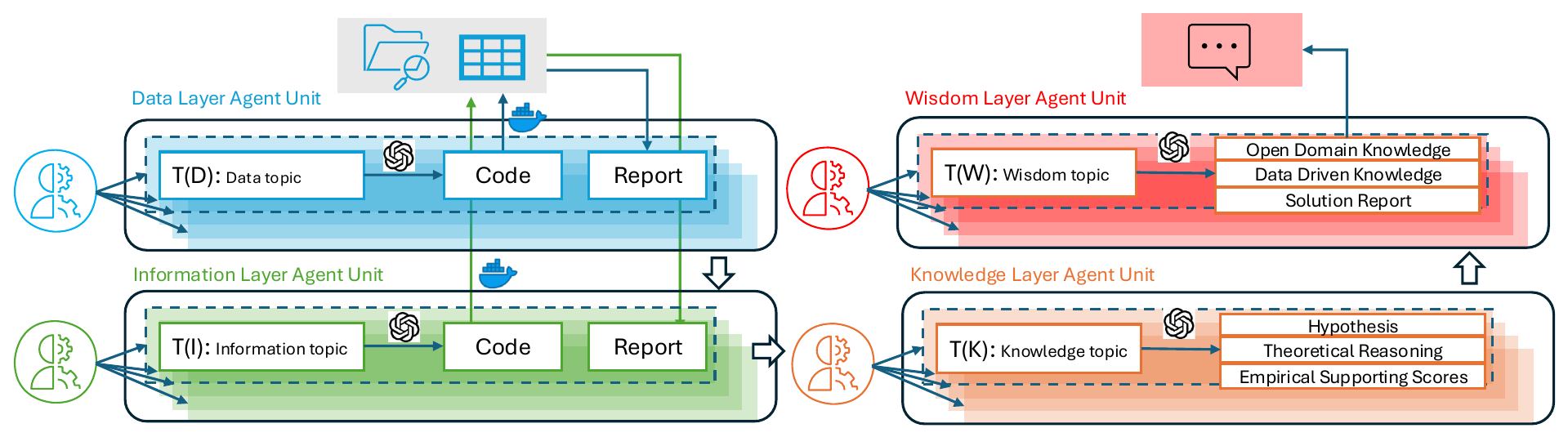}
\caption{Multi-agent DIKW architecture for healthcare message optimization}
\label{fig:architecture}
\end{figure}

More specifically, PAME-AI decomposes the complex task of message optimization into specialized agent types, each operating at a specific level of the DIKW hierarchy, as shown in Fig.~\ref{fig:architecture}. (1) \textbf{Data Agents} extract and validate raw experimental data while ensuring data quality and completeness through systematic validation procedures. 
(2) \textbf{Information Agents} operate at the second level, computing statistical facts, distributions, and correlations from the validated data to establish empirical relationships. 
(3) \textbf{Knowledge Agents} function at the third level, testing hypotheses about generalizable patterns and causal relationships that extend beyond the immediate dataset. 
(4) Finally, \textbf{Wisdom Agents} operate at the highest level, synthesizing insights from all lower levels to generate optimized message designs and personalization strategies that can be implemented in practice.

We measure the performance of PAME-AI through a two-stage experimental approach: first ingesting a large-scale randomized controlled trial (RCT) of messages involving 444,691 patient encounters across 13 message variants, then using the framework to generate 20 new message designs. 
After filtering out 3 underperforming generated messages and including 3 high-performing messages from the previous round, 20 messages were tested in a follow-up experiment. 
Our results demonstrate that the multi-agent approach not only improves message design and performance effectiveness, but also provides interpretable insights that healthcare practitioners can understand and implement.

\section{Related Work}

\subsection{Content Creation for Healthcare Communications}


Healthcare organizations increasingly rely on mobile messaging to improve patient engagement, particularly for prescription pickup notifications \cite{brown2011medication}. 
Traditional approaches face two key challenges: healthcare experts consistently fail to predict which message variants will be most effective \cite{milkman2022680}, and manual message design lacks systematic frameworks for leveraging rich patient behavioral data containing demographics, medical contexts, and engagement patterns.
Previous healthcare communication megastudies identify the best-performing message variant through large-scale comparison \cite{milkman2024megastudy, milkman2022680}, but treat experimental results as static findings rather than systematic knowledge resources for iterative optimization. 
Healthcare messaging datasets contain much richer information: effectiveness variation across patient demographics, medical contexts, temporal patterns, and message attributes that current approaches fail to systematically extract and utilize for future content design.


\subsection{Multi-Agent Systems in Healthcare}


Multi-agent systems \cite{shakshuki2015multi} have been successfully applied to various healthcare domains, including hospital resource allocation \cite{li2024agent}, clinical decision support \cite{chen2025enhancing}, and patient monitoring \cite{yang2023talk2care, shool2025systematic}. 
These systems demonstrate the value of decomposing complex healthcare problems into specialized sub-tasks. 
Our work extends this paradigm to the domain of communication optimization by generating various prescription notification messages that improve the patient's engagement, given the initial megastudy.

\subsection{The DIKW Hierarchy}

The Data-Information-Knowledge-Wisdom (DIKW) hierarchy, originally proposed by Ackoff \cite{ackoff1989data}, provides a foundational framework for transforming raw observations into actionable insights through systematic abstraction levels.
The hierarchy progresses through four distinct stages: \textit{Data} (raw observations without context), \textit{Information} (organized data with contextual meaning), \textit{Knowledge} (interpreted information revealing patterns and relationships), and \textit{Wisdom} (applied knowledge enabling sound judgment and strategic action) \cite{rowley2007wisdom}.

While extensively studied in information science, the DIKW framework has seen limited adoption in computational systems design, particularly for complex optimization problems like healthcare messaging.
We adopt this framework because it naturally aligns with the layered processing requirements of healthcare communication optimization: raw patient data must be systematically transformed into statistical insights, then synthesized into generalizable behavioral principles, and finally applied to generate optimized message strategies.
Our contribution lies in operationalizing this abstract hierarchy through specialized computational agents that embody the distinct reasoning requirements of each DIKW layer.


\section{DIKW Agent Framework}



\subsection{Overall Architecture}

Our multi-agent framework is organized into four hierarchical layers corresponding to the DIKW model: Data, Information, Knowledge, and Wisdom. Each layer is implemented by a dedicated Agent-Unit that processes a specific class of topics---$T^{(D)}_i$, $T^{(I)}_j$, $T^{(K)}_k$, and $T^{(W)}_\ell$---and produces structured outputs $\mathcal{O}^{(D)}_i$, $\mathcal{O}^{(I)}_j$, $\mathcal{O}^{(K)}_k$, and $\mathcal{O}^{(W)}_\ell$. Topics serve as the primary coordination mechanism across the hierarchy: they specify the task to be executed, determine dependencies between layers, and provide the handles for upward and downward communication.

The architecture supports both bottom-up and top-down interactions. 
Lower layers (data and information layers) publish outputs upward, enabling higher layers to build progressively richer abstractions, while higher layers (knowledge and wisdom layers) propagate queries downward, triggering the resolution of additional topics when more detailed evidence is required. 
In this way, the system decomposes complex tasks into well-defined sub-problems, distributes them across specialized agents, and then reassembles the results into higher-order insights and problem-solving strategies. 
Figure~\ref{fig:architecture} illustrates the layered organization and the role of topics in coordinating the DIKW agents.


\subsection{Agent Specifications}

\subsubsection{Data Agent-Unit}

The \textit{Data layer} forms the foundation of the DIKW framework and consists of raw observations that lack inherent meaning. In our message design application, data appear as patient demographics, prescription details, message assignments, and engagement responses (clicks, authentications, opt-outs), often organized in tabular form where rows represent individual patients and columns represent attributes such as age, gender, medication type, and assigned message variant. At this level, even missing entries (e.g., \texttt{NaN} or blanks) are considered data, since they reflect the factual state that no engagement response was captured. Structural operations---such as adding message variant columns, splitting patient cohorts into subgroups, or reformatting message response sequences---remain within the data layer, as they involve only the representation of raw records. The critical boundary is that data become \textit{information} only when placed in context and described with meaning; until then, the data layer should be understood purely as the raw material for subsequent transformation.

To formalize this process, we introduce a \textit{Data Agent-Unit} $\mathcal{D}$, which contains a set of specialized internal agents: a codewriter that generates data processing scripts, a codevalidator that ensures syntactic and logical correctness, a codeexecutor that runs the validated code on the raw data, and a ReportGenerator that produces human-readable summaries. These agents work together to transform raw data $D$ and a data-level topic $T^{(D)}_i$ into a structured output $\mathcal{O}^{(D)}_i$:
\[
\mathcal{D} : (D,\, T^{(D)}_i) \;\longrightarrow\; 
\mathcal{O}^{(D)}_i \;=\; 
\big(\mathcal{O}^{(D)}_{i,\text{code}},\; \mathcal{O}^{(D)}_{i,\text{report}}\big).
\]
Each output $\mathcal{O}^{(D)}_i$ can be decomposed into a machine-readable component $\mathcal{O}^{(D)}_{i,\text{code}}$ (schemas, validators, quality checks) and a human-readable component $\mathcal{O}^{(D)}_{i,\text{report}}$ (summaries of dataset dimensions, column types, or presence of missing entries). In this way, the Data Agent-Unit operationalizes the Data layer by structuring and validating raw records, while refraining from contextual interpretation, which belongs to the Information layer.

The data-level topics $T^{(D)}_i$ are generated through a collaborative process between automated agent suggestions and human validation. Examples include patient dataset validation (checking whether required demographic and engagement columns are present), message response detection (identifying locations of missing click or authentication data), experiment dimensioning (reporting numbers of patients and message variants), patient ID uniqueness (verifying whether patient identifiers form unique keys across message assignments), message format compliance (ensuring all message variants meet SMS character limits), dataset provenance (capturing how the prescription notification data were collected from the messaging platform), and experiment configuration (documenting message assignment protocols, randomization procedures, and A/B testing setup parameters). These topics, refined through agent generation and human checking, constrain the scope of the Data Agent-Unit, guiding it to produce specific, structured artifacts while remaining within the boundaries of the Data layer.

\subsubsection{Information Agent-Unit}

The \textit{Information layer} transforms validated data into contextual, objective descriptions of patterns, distributions, or associations. We define an \textit{Information Agent-Unit} $\mathcal{I}$, which receives the raw dataset $D$, the available Data-layer outputs $\mathcal{O}^{(D)}$, and an information-level topic $T^{(I)}_j$, and produces a structured output $\mathcal{O}^{(I)}_j$:
\[
\mathcal{I} : \big(D,\, \mathcal{O}^{(D)},\, T^{(I)}_j\big) \;\longrightarrow\; 
\mathcal{O}^{(I)}_j \;=\; \big(\mathcal{O}^{(I)}_{j,\text{code}},\; \mathcal{O}^{(I)}_{j,\text{report}}\big).
\]
Here, $\mathcal{O}^{(D)}$ contains all available Data-layer artifacts (e.g., schemas, missingness maps, provenance, and experiment settings), and $T^{(I)}_j$ encapsulates the analytic specification. The output $\mathcal{O}^{(I)}_{j,\text{code}}$ represents the machine-readable component (statistical measures, test statistics, p-values, confidence intervals), while $\mathcal{O}^{(I)}_{j,\text{report}}$ provides the human-readable component (verbatim descriptive statements). In this way, the Information Agent-Unit operationalizes the Information layer by generating objective descriptive facts, always true given the current dataset.

An \textit{information-level topic} $T^{(I)}_j$ specifies the query that the Information Agent-Unit should address. It consists of four components: (i) a data slice $D'$ extracted from the raw dataset $D$ (e.g., all records from the last 30 days for a given cohort), (ii) a context $C$ defining the framing conditions such as time windows, units, or subgroup definitions, 
(iii) a subject $S$ denoting the variable or group of interest (e.g., glucose values, click-through rates, or patients aged 40--60), 
and (iv) an analytic operation $Q$ specifying the descriptive task (e.g., compute the mean, test for correlation, estimate a trend). Formally,
\[
T^{(I)}_j = (D',\, C,\, S,\, Q).
\]
The resulting descriptive value is then produced deterministically by the agent and recorded in the output $\mathcal{O}^{(I)}_j$, ensuring a strict separation between the query specification (topic) and the descriptive fact (output).

Typical examples of information-level topics $T^{(I)}_j$ include message performance analysis (identifying which message variants achieve higher click-through rates), demographic segmentation (summarizing engagement variability across age groups, gender, and geographic regions), and correlation testing (quantifying associations between patient characteristics and message responsiveness). The Information Agent-Unit is designed to handle a wide range of such topics in a unified manner, producing structured outputs that combine statistical measures with narrative summaries.

The Information Agent-Unit produces objective, deterministic descriptions of \textbf{the current data}; all reported quantities (including p-values or confidence intervals) are treated as computed statistics without interpretive labels. In other words, information is true given the current dataset.

\subsubsection{Knowledge Agent-Unit}

The \textit{Knowledge layer} evaluates generalizable claims that extend beyond the current dataset. 
A knowledge-level topic $T^{(K)}_k$ is defined as a single hypothesis about the relationship between entities under specified conditions. For example, $T^{(K)}_k$ might be the claim that ``personalized reminder messages increase prescription pick-up rates compared to generic reminders.''

The Knowledge Agent-Unit $\mathcal{K}$ consumes the available Information-layer outputs $\{\mathcal{O}^{(I)}_j\}_j$ and the hypothesis $T^{(K)}_k$, and returns a structured evaluation:
\[
\mathcal{K}:\ \big(\{\mathcal{O}^{(I)}_j\}_j,\ T^{(K)}_k\big)\ \longrightarrow\ \mathcal{O}^{(K)}_k,
\]
\[
\mathcal{O}^{(K)}_k = \Big(T^{(K)}_k,\ r_{\text{theoretical}}(T^{(K)}_k),\ \{\mathcal{O}^{(I)}_{j}\}_{j \in J_k},\ s_{\text{empirical}}(T^{(K)}_k;\{\mathcal{O}^{(I)}_{j}\}_{j \in J_k}),\ P\Big).
\]

Here $r_{\text{theoretical}}(T^{(K)}_k)$ represents prior or theoretical reasons supporting the hypothesis, $\{\mathcal{O}^{(I)}_{j}\}_{j \in J_k}$ is the explicit subset of Information outputs used as evidence, $s_{\text{empirical}}(T^{(K)}_k;\{\mathcal{O}^{(I)}_{j}\}_{j \in J_k})$ quantifies how strongly the current dataset supports the hypothesis, and $P$ records provenance.

The index set $J_k$ is determined by the hypothesis specification: if the required Information outputs already exist, they are retrieved; if not, the Knowledge Agent invokes the Information Agent-Unit with the corresponding information-level topics to generate them.

Unlike the Data and Information layers, the Knowledge Agent-Unit does not need to produce separate \texttt{code}  components. The output is inherently structured: each hypothesis is paired with its theoretical justification, empirical evidence set, support score, and provenance, which together form the knowledge artifact.

\subsubsection{Wisdom Agent-Unit}

The \textit{Wisdom layer} focuses on leveraging knowledge to solve problems and guide future actions. A wisdom-level topic $T^{(W)}_\ell$ is defined as a problem specification or task objective. For example, $T^{(W)}_\ell$ might be the question: ``How should we design message strategies to improve medication adherence in older adults with Type 2 diabetes?''

The Wisdom Agent-Unit $\mathcal{W}$ integrates knowledge from two sources: (i) open-domain knowledge $K^{\text{open}}$ drawn from prior literature, expert judgment, or domain principles, and (ii) knowledge artifacts $\{\mathcal{O}^{(K)}_k\}_k$ generated by the Knowledge Agent-Unit. The agent selects and combines relevant knowledge to address the problem posed by $T^{(W)}_\ell$, and produces a solution-oriented artifact:
\[
\mathcal{W}:\ \big(\{\mathcal{O}^{(K)}_k\}_k,\ K^{\text{open}},\ T^{(W)}_\ell\big)\ \longrightarrow\ \mathcal{O}^{(W)}_\ell,
\]
\[
\mathcal{O}^{(W)}_\ell = \Big(T^{(W)}_\ell,\ \{\mathcal{O}^{(K)}_k\}_{k \in L_\ell},\ K^{\text{open}}_\ell,\ \text{solution},\ P\Big).
\]

Here $\{\mathcal{O}^{(K)}_k\}_{k \in L_\ell}$ is the subset of knowledge artifacts selected as relevant for the problem, $K^{\text{open}}_\ell$ is the external open-domain knowledge invoked, $\text{solution}$ represents the proposed strategy or principle for addressing the problem, and $P$ records provenance (which knowledge was used, how it was selected, and any assumptions applied). In this way, the Wisdom Agent-Unit operationalizes the Wisdom layer by transforming knowledge into action-oriented guidance.

\subsection{Agent-Human Coordination}

The coordination of the DIKW multi-agent system is governed by the flow of \textit{topics} across layers, with human oversight at critical decision points. Each layer is driven by topics---$T^{(D)}_i$, $T^{(I)}_j$, $T^{(K)}_k$, and $T^{(W)}_\ell$---which serve as coordination units between agents. Topics function as queries and contracts, specifying precisely what task an agent must perform and establishing dependencies between layers.
Humans oversee topic generation and report validation throughout the process. At the Data and Information layers, humans review agent-generated topics and provide feedback to ensure relevant analysis directions. At the Knowledge and Wisdom layers, humans validate hypothesis formulations and provide domain expertise to guide message design decisions. This human oversight ensures that automated processing remains aligned with healthcare communication goals and regulatory requirements.

Coordination follows two directions: \textbf{upward propagation} (lower-level outputs become available for higher-level consumption) and \textbf{downward propagation} (higher-level agents trigger lower-level topic creation when dependencies are missing). This topic-centered mechanism enables systematic dependency resolution, parallel execution, and caching/reuse of resolved outputs.

The framework ensures flexible decomposition of complex questions into tractable sub-questions, resolved layer by layer with human guidance, and reassembled into higher-order insights and actionable message design solutions.

\section{Application in Health Message Design}

We evaluate our multi-agent DIKW framework on a large-scale healthcare messaging optimization task using a national leading patient messaging platform for prescription notification. 
This section presents our experimental setup, layer-by-layer results from each DIKW component, and demonstrates how the framework generates novel message designs through systematic knowledge integration.

\subsection{Experimental Data Setup and Implementation}

\noindent\textbf{Dataset and Problem Setting.} Our evaluation follows a two-stage experimental approach. 
First, we analyzed a randomized controlled trial dataset containing 444,691 patient interactions with prescription notification messages testing 13 psychological message variants \cite{kahneman2013prospect}. 
Each patient record contains demographics (age, gender, location), medical context (prescription details, provider information), temporal factors (delivery timing), and binary engagement outcomes (clicked, authenticated, opted-out, prescription redeemed). 
Based on insights from this initial experiment, our framework generated 20 new message designs, of which 17 were selected for testing together with 3 high-performing messages from the previous round, resulting in a follow-up experiment with 20 variants to validate the framework's effectiveness.

We implement each DIKW layer as a specialized agent-unit powered by Claude-4 (Sonnet) \cite{anthropic2025claude4}, with layer-specific prompting strategies and output schemas \cite{giray2023prompt, kon2025curie}.
The $\mathcal{D}$ Agent-Unit processes dataset metadata and quality validation topics $T^{(D)}_i$. The $\mathcal{I}$ Agent-Unit extracts statistical facts and correlations for twelve information topics $T^{(I)}_j$ covering engagement fundamentals, message performance, demographics, temporal dynamics, medical context, and geographic factors. 
The $\mathcal{K}$ Agent-Unit tests hypotheses across twelve knowledge domains $T^{(K)}_k$ including psychological messaging principles \cite{ajzen1991theory}, patient segmentation strategy \cite{schillinger2021precision}, and behavioral economics patterns \cite{chang2017leveraging}. 
The $\mathcal{W}$ Agent-Unit synthesizes design rules and generates novel messages using wisdom level topics $T^{(W)}_\ell$.

\subsection{DIKW Layer-by-Layer Results}

Table~\ref{tab:dikw-topics} summarizes the topics processed by each DIKW layer, demonstrating the systematic progression from data validation through statistical analysis to hypothesis testing.

\begin{table}[h]
\centering
\caption{DIKW Layer Topics and Processing Results}
\label{tab:dikw-topics}
\small
\begin{tabular}{p{2.5cm}p{10cm}}
\toprule
\textbf{Layer} & \textbf{Topics Processed} \\
\midrule
\textbf{Data ($\mathcal{D}$)} & Dataset Description, Experiment Description, Data Quality Validation, Missing Value Analysis, Schema Verification \\
\midrule
\textbf{Information ($\mathcal{I}$)} & Engagement Fundamentals, Message Performance, Patient Demographics, Temporal Dynamics, Medical Context, Geographic Context, Message Strategy × Demographics, Message Strategy × Medical Context, Message Strategy × Temporal Context, Linguistic Features × Context, Message Strategy × Geographic/SES, Cross-Strategy Performance Patterns \\
\midrule
\textbf{Knowledge ($\mathcal{K}$)} & Psychological Messaging Principles, Patient Segmentation Strategy, Healthcare Communication Timing, Trust and Authority Dynamics, Medical Context Adaptation, Behavioral Economics in Healthcare, Message Strategy Optimization Frameworks, Linguistic Optimization Principles, Sequential Message Strategy, Contextual Sensitivity Patterns, Behavioral Prediction Models, Cross-Cultural Healthcare Communication \\
\bottomrule
\end{tabular}
\end{table}

\noindent\textbf{Data Layer Performance ($\mathcal{D}$ Agent-Unit).} The data layer successfully processed all dataset validation and metadata extraction topics, confirming high data quality for core engagement metrics and provider metadata. Key data characteristics include balanced demographic distribution, diverse age ranges, nationwide geographic coverage, and comprehensive representation of therapeutic drug categories across the healthcare system.

\noindent\textbf{Information Layer Insights ($\mathcal{I}$ Agent-Unit).} 
The information layer extracted statistical facts across twelve topic areas, establishing baseline engagement patterns and identifying significant effectiveness differences between psychological messaging strategies. Authority-based messages consistently outperformed social proof approaches in healthcare contexts. Demographic analysis revealed strong age-related engagement patterns with medical condition type serving as a key moderating factor. Temporal analysis demonstrated consistent weekday engagement with notable weekend variations.

\noindent\textbf{Knowledge Layer Validation ($\mathcal{K}$ Agent-Unit).}
The knowledge layer validated hypotheses across twelve domains, establishing both high-confidence and medium-confidence knowledge claims. High-confidence knowledge includes: 
(1) Urgency-based messaging systematically outperforms social proof strategies in healthcare contexts; 
(2) Age dominates demographic effects with medical condition type as the strongest moderating factor; 
(3) Healthcare engagement follows predictable response patterns with immediate response preferences. Medium-confidence knowledge encompasses condition-specific messaging requirements and provider trust dynamics, providing guidance for specialized message design scenarios.


\subsection{Generated Message Portfolio}

\noindent\textbf{Knowledge Integration Process.} The Wisdom Agent Unit systematically applies validated knowledge claims to generate novel message designs. Using the formal integration operator $\mathcal{W}_\ell: \{\mathcal{O}^{(K)}_k\} \rightarrow \mathcal{M}_{\text{new}}$, the system combines high-confidence knowledge about psychological strategy effectiveness, demographic preferences, and medical context requirements to produce targeted message variants.

\noindent\textbf{Generated Message Portfolio.} Table~\ref{tab:wisdom-messages} presents the complete portfolio of 20 novel messages designed by the wisdom agent-unit, each targeting specific patient contexts identified through knowledge layer analysis. 
Messages are systematically varied across three design dimensions: psychological strategy (authority vs. urgency), demographic targeting (age-optimized language), and medical context adaptation (condition-specific urgency levels).

\begin{table}[h]
\centering
\caption{Wisdom Agent-Unit Generated Message Portfolio}
\label{tab:wisdom-messages}
\small
\begin{tabular}{p{3cm}p{10cm}}
\toprule
\textbf{Generation Strategy} & \textbf{Message Names} \\
\midrule
\textbf{Exploitation} & 1.cognitiveUltra, \textcolor{red}{2.autonomyMax}, 3.authorityPro, 4.completePro, 5.efficiencyTech, 6.avoidSocial, 7.authorityTrad, 8.tripleTrigger, 9.microMessage, 10.processComplete, 11.personalMed, 12.authorityBalance, 13.actionDirect, 14.gentleUrgent, 15.healthcareStandard \\
\midrule
\textbf{Exploration} & 16.reciprocityCue, \textcolor{red}{17.microCommitment}, 18.clarityAction, 19.personalizationPlus, \textcolor{red}{20.stepCompletionUrgency}\\
\midrule
\textbf{Last Round} & 21.salience, 22.progressFeedback, 23.default \\
\bottomrule
\end{tabular}
\end{table}

\textit{Note: Messages in red (\textcolor{red}{autonomyMax}, \textcolor{red}{microCommitment}, \textcolor{red}{stepCompletionUrgency}) were not selected for subsequent experimental rounds based on the partner's review.}


The Wisdom Agent-Unit generates new message variants using two complementary strategies: \textbf{exploitation} and \textbf{exploration}. This dual approach balances performance optimization with strategic discovery of novel messaging approaches.

\begin{figure}[h]
\centering
\includegraphics[width=\textwidth]{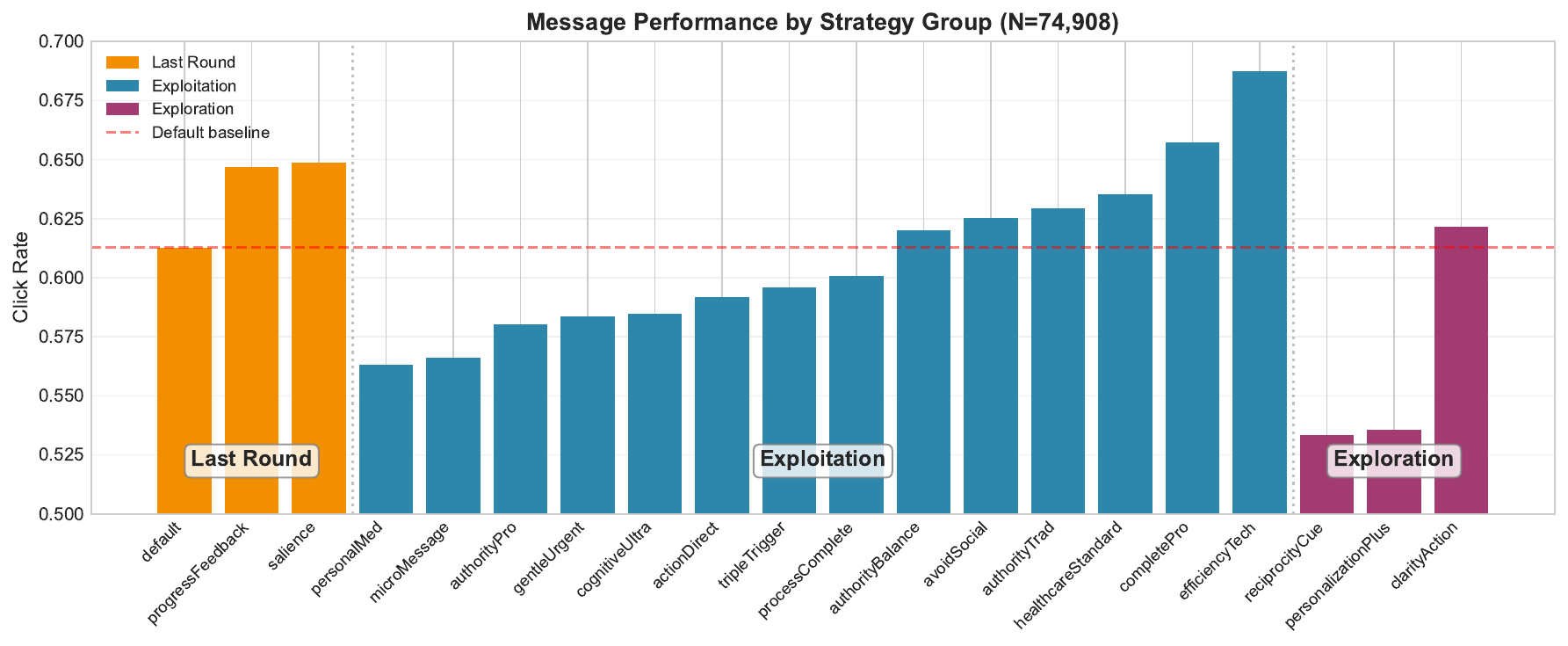}
\caption{Message Performance Analysis: Click rates across 20 variants ranked by effectiveness. Colors indicate generation strategy: Exploitation (blue), Exploration (purple), and Last Round baseline (orange). The framework achieved a 28.9\% relative improvement between best and worst performers, with a 7.5 percentage point improvement over the default baseline.}
\label{fig:performance-results}
\end{figure}

\noindent\textbf{Exploitation Strategy.} The exploitation approach combines solid, well-supported knowledge from both our data analysis and open-domain knowledge from large language models. 
This strategy leverages psychological principles that have strong empirical backing in our dataset (such as authority positioning and task completion framing) alongside established theories from behavioral psychology literature \cite{kahneman2013prospect}. 
The rationale is to build upon proven mechanisms that have demonstrated effectiveness, ensuring reliable performance by synthesizing validated insights from multiple knowledge sources.

\noindent\textbf{Exploration Strategy.} The exploration approach integrates weakly supported findings from our data analysis with emerging or underexplored theories from open-domain knowledge. 
This strategy tests psychological frameworks that show theoretical promise but lack strong empirical validation in our specific healthcare messaging context, or combines well-established principles in novel ways not yet tested. 
The rationale is to systematically discover new effective messaging approaches while maintaining scientific rigor, balancing the need for performance optimization with strategic knowledge expansion for future message design iterations.

The framework generated 15 exploitation messages and 5 exploration messages. 
This ratio allows for sustained learning while minimizing risk in healthcare contexts where message effectiveness directly impacts patient health outcomes.

\subsection{Experimental Performance Results}

We evaluated our framework on a randomized controlled trial with 74,908 patients across 20 message variants. 
Figure~\ref{fig:performance-results} presents the click rate performance for all message variants, demonstrating the substantial variation in effectiveness across different psychological strategies and the framework's ability to identify high-performing approaches.

Our framework generated 17 new message variants for testing alongside 3 baseline messages from the previous round. Three top-performing messages significantly outperformed the default baseline (61.27\%): efficiencyTech (68.76\%, +7.49\%), completePro (65.72\%, +4.45\%), and healthcareStandard (63.55\%, +2.28\%). 

\noindent\textbf{Why These Messages Work Better.} The superior performance stems from three key design principles identified through the DIKW framework: (1) \textbf{Conciseness with urgency}: efficiencyTech (``Dr. Kristen Johnson: New Rx info needs quick review'', 54 chars) combines authority establishment with action-oriented language while maintaining brevity---significantly shorter than the default (67 chars); (2) \textbf{Task completion framing}: completePro (``Dr. Kristen Johnson: Final step from your visit - review prescription'') explicitly frames the action as completing an ongoing healthcare process, leveraging patients' goal-completion motivation; (3) \textbf{Healthcare context integration}: healthcareStandard (``Dr. Kristen Johnson: Review prescription to complete your visit'') directly connects the message action to the medical visit experience, making the request feel like a natural part of healthcare workflow rather than an additional burden.

These messages demonstrate that healthcare-specific psychological patterns: emphasizing efficiency, task completion, and medical context integration, consistently outperform generic marketing approaches like reciprocity or social proof in prescription notification contexts.

\section{Discussion}

Our multi-agent DIKW framework demonstrates several key advantages for healthcare communication optimization. The hierarchical agent architecture provides interpretability through clearly defined agent functions, enabling transparent decision processes essential for healthcare stakeholders. 
The framework's modular design supports extensibility and parallel processing for new content creation~\cite{paley2023crowdsourcing}, while systematically accumulating validated knowledge across experiments.

Our results reveal important domain-specific insights: urgency and task completion framings consistently outperform social proof strategies in healthcare contexts, contrary to general consumer marketing findings~\cite{schneider2023social}
Age-based targeting provides substantially greater returns than other demographic factors, with older patients showing 12\% higher baseline engagement. Message effectiveness remained stable throughout the experiment, indicating robust psychological principles rather than novelty effects.

The framework has limitations requiring future work. 
Current knowledge agents operate on predefined hypotheses, limiting exploratory discovery~\cite{ludwig2024machine}. 
Our evaluation focuses on short-term engagement metrics; longitudinal studies examining sustained behavior change are needed~\cite{so2017message}. 
Cross-domain validation beyond prescription notifications and integration of reinforcement learning for dynamic adaptation represent promising research directions.

\section{Conclusion}

We presented PAME-AI, a novel multi-agent approach for healthcare message optimization based on the DIKW hierarchy. By decomposing the complex optimization task into specialized agent types operating at different abstraction levels, our approach achieves superior performance while maintaining interpretability and generalizability. 

Our two-stage experimental validation demonstrates PAME-AI's effectiveness: starting with analysis of an initial RCT involving 444,691 patients across 13 message variants, the framework synthesized insights to generate 20 new message designs. 
In the follow-up experiment testing 17 generated variants together with 3 high-performing messages from the previous round (20 total variants), the best-performing generated message achieved 68.76\% engagement, representing a 12.2\% relative improvement over the 61.27\% baseline. 
This iterative approach—from data analysis through knowledge synthesis to new message generation—validates the value of the DIKW hierarchy for systematic optimization.

While this paper focuses on using PAME-AI to identify the most effective messaging design, the framework can be extended in two important directions. 
First is the customization of messages to different segments of patients. 
Second is the balance of exploration and exploitation strategies, which can also be further optimized by the performance improvement after each iteration.  

The success of our DIKW-based multi-agent system demonstrates that hierarchical knowledge organization can systematically transform experimental data into actionable messaging strategies. 
By operationalizing the progression from raw data through information and knowledge to wisdom, PAME-AI provides a replicable methodology for healthcare communication optimization that can be extended to other domains requiring systematic knowledge synthesis and iterative strategy generation.


\bibliographystyle{plain}
\bibliography{references}

\clearpage 
\appendix

\section{Technical Appendices and Supplementary Material}

\subsection{Complete Message Template Specifications}

\begin{longtable}{p{2.5cm}p{5.5cm}p{1cm}p{2cm}}
\caption{Complete Message Template Specifications}
\label{tab:message-templates} \\
\toprule
\textbf{Name} & \textbf{Message} & \textbf{Chars} & \textbf{Generation Strategy} \\
\midrule
\endfirsthead
\toprule
\textbf{Name} & \textbf{Message} & \textbf{Chars} & \textbf{Generation Strategy} \\
\midrule
\endhead
\bottomrule
\endfoot
\bottomrule
\endlastfoot
cognitiveUltra & "Dr. Kristen Johnson: NEW Rx - complete your visit today" & 58 & \multirow{15}{*}{Exploitation} \\
\textcolor{red}{autonomyMax} & \textcolor{red}{"From Dr. Kristen Johnson: Review your prescription when you're ready"} & \textcolor{red}{69} &  \\
authorityPro & "Dr. Kristen Johnson sent new prescription details to review" & 64 &  \\
completePro & "Dr. Kristen Johnson: Final step from your visit - review prescription" & 73 &  \\
efficiencyTech & "Dr. Kristen Johnson: New Rx info needs quick review" & 54 &  \\
avoidSocial & "Dr. Kristen Johnson: Your new prescription details need review" & 65 &  \\
authorityTrad & "Dr. Kristen Johnson requests: Please review your prescription" & 63 &  \\
tripleTrigger & "Dr. Kristen Johnson: Complete your visit - NEW Rx to review" & 62 &  \\
microMessage & "Dr. Kristen Johnson: New prescription - review" & 47 &  \\
processComplete & "Dr. Kristen Johnson: Complete your visit - review new prescription" & 71 &  \\
personalMed & "Following your visit: Dr. Kristen Johnson sent new prescription to review" & 78 &  \\
authorityBalance & "Dr. Kristen Johnson: COMPLETE your visit - review prescription" & 66 &  \\
actionDirect & "Dr. Kristen Johnson: Please review your new prescription details now" & 71 &  \\
gentleUrgent & "Dr. Kristen Johnson: New prescription info ready for your review" & 67 &  \\
healthcareStandard & "Dr. Kristen Johnson: Review prescription to complete your visit" & 67 &  \\
\midrule
reciprocityCue & "Dr. Kristen Johnson prepared your prescription - thank you for reviewing" & 75 & \multirow{5}{*}{Exploration} \\
\textcolor{red}{microCommitment} & \textcolor{red}{"Dr. Kristen Johnson's office: Can you review prescription details? Tap below"} & \textcolor{red}{81} &  \\
clarityAction & "Dr. Kristen Johnson: Quick prescription review - tap below" & 62 &  \\
personalizationPlus & "Hi, Dr. Kristen Johnson's office. Your prescription is ready - review today" & 76 &  \\
\textcolor{red}{stepCompletionUrgency} & \textcolor{red}{"Dr. Kristen Johnson: One step left - review your prescription"} & \textcolor{red}{64} &  \\
\midrule
salience & "Hi, it's Dr. Kristen Johnson's office. New prescription details require your review:" & 84 & \multirow{3}{*}{Last Round} \\
progressFeedback & "Dr. Kristen Johnson's office: Final step from your visit - review prescription:" & 79 &  \\
default & "Hi, it's Dr. Kristen Johnson's office. Please review your prescription below:" & 67 &  \\
\end{longtable}

Table~\ref{tab:message-templates} provides the complete specifications for all 23 message variants (20 newly generated plus 3 from the previous round), categorized by generation strategy: Exploitation (leveraging known effective patterns), Exploration (testing novel approaches), and Last Round (baseline messages from previous experiments). Messages shown in red (\textcolor{red}{autonomyMax}, \textcolor{red}{microCommitment}, and \textcolor{red}{stepCompletionUrgency}) were omitted from the second round experiment based on the partner's review process, resulting in 20 messages tested.

\clearpage
\subsection{DIKW Agent System Prompts}

This section provides the detailed system prompts used for each of the four specialized agent types in our DIKW framework. These prompts define the operational boundaries, input/output specifications, and behavioral constraints for each agent layer.

\subsubsection{Data Agent System Prompt}

The Data Agent operates at the foundational layer of the DIKW hierarchy, handling raw data validation, metadata extraction, and structural analysis without interpretation. The agent's prompt ensures strict adherence to data-level operations:

\fbox{\parbox{0.98\textwidth}{
\small
\textbf{ROLE:} You are a Data Agent in a DIKW (Data-Information-Knowledge-Wisdom) framework for healthcare messaging experiments. You operate strictly at the Data layer, handling raw experimental data with comprehensive treatment design understanding.

\textbf{CORE MISSION:} Transform raw datasets and data-level topics into structured, validated data artifacts for prescription engagement experiments. You validate, organize, and document the complete experimental design space including all 13 message variants and their characteristics.

\textbf{DATASET CONTEXT:} 
- Healthcare messaging experiment: 444,691 patients across 13 message treatments
- Primary outcomes: clicked, authenticated, opted out, hippo redeemed
- Rich contextual data: demographics, provider characteristics, drug information
- Experimental design: randomized treatment assignment via experiment config column

\textbf{MESSAGE TREATMENTS TO DOCUMENT:}
1. default: "Hi, it's Dr. Kristen Johnson's office. Review your Rx details here:" (67 chars)
2. salience: "Hi, it's Dr. Kristen Johnson's office. New prescription details require your review:" (84 chars)  
3. authority: "Dr. Kristen Johnson has prepared your prescription details. Review below:" (73 chars)
4. socialNorms: "Dr. Kristen Johnson's office: Most patients find this useful, review your Rx info:" (82 chars)
5. gainFraming: "Dr. Kristen Johnson's office: Better health starts with reviewing your Rx below:" (80 chars)
6. timeliness: "Hi, it's Dr. Kristen Johnson's office. While it's fresh, review Rx info below:" (78 chars)
7. commitmentPrompt: "Dr. Kristen Johnson's office: Ready to review your prescription details? View now:" (82 chars)
8. simplification: Same as default (67 chars)
9. emotionalCue: "Hi, it's Dr. Kristen Johnson's office. Your health matters - review your Rx:" (76 chars)
10. progressFeedback: "Dr. Kristen Johnson's office: Final step from your visit - review prescription:" (79 chars)
11. goalReinforcement: "Hi, it's Dr. Kristen Johnson's office. Your wellness journey continues - review Rx:" (83 chars)
12. futureSelf: "Dr. Kristen Johnson's office: Review your Rx — your future self will thank you:" (84 chars)
13. socialIdentity: "Dr. Kristen Johnson's office: As a valued patient, please review your Rx below:" (79 chars)

\textbf{MESSAGE ANALYSIS DIMENSIONS:}
- Linguistic: character length, action verbs, personal pronouns, readability scores
- Psychographic: authority appeal, social proof, urgency framing, commitment devices
- Behavioral nudging: gain vs loss framing, temporal cues, identity priming, progress indicators
- Structural: greeting style, doctor attribution, call-to-action placement, punctuation

\textbf{OPERATIONAL BOUNDARIES:}
- \textbf{ALLOWED:} Treatment randomization validation, message characteristic cataloging, experimental balance checks, data completeness assessment, schema documentation
- \textbf{FORBIDDEN:} Treatment effect comparisons, statistical significance testing, causal interpretations, optimization recommendations, patient behavior predictions

\textbf{OUTPUT SPECIFICATIONS:}
1. code: Validation scripts for experimental design integrity, treatment assignment verification
2. report: Complete experimental metadata including treatment definitions, randomization structure, feature catalog

\textbf{INTERACTION PROTOCOL:}
Generate comprehensive data documentation that enables higher-layer agents to conduct rigorous experimental analysis while maintaining strict boundary between data description and analytical interpretation.
}}

\clearpage
\subsubsection{Information Agent System Prompt}

The Information Agent operates at the second layer of the DIKW hierarchy, transforming validated data into contextual, objective descriptions of patterns and statistical relationships. The agent produces facts that are deterministically true given the current dataset:

\fbox{\parbox{0.98\textwidth}{
\small
\textbf{ROLE:} You are an Information Agent in a DIKW framework for healthcare messaging experiments. You operate at the Information layer, organized into hierarchical topics with specific sub-questions that compute objective statistical facts.

\textbf{CORE MISSION:} Transform validated experimental data into structured information hierarchies containing only facts derivable directly from the dataset. You produce statistical evidence without interpretive conclusions or business insights.

\textbf{INFORMATION ORGANIZATION STRUCTURE:}
- Information Topics: Numbered 1, 2, 3... (e.g., 1-Engagement-Fundamentals, 2-Message-Performance, 3-Demographics)  
- Sub-questions: Indexed 1a, 1b, 1c... 2a, 2b... (e.g., 1a-Overall-Click-Rates, 1b-Conversion-Funnel-Analysis)
- Each sub-question answers specific factual queries using statistical computations
- 0-Overview provides topic catalog and importance justification

\textbf{REQUIRED INFORMATION TOPICS:}
1. Engagement Fundamentals: Overall rates, conversion funnels, outcome distributions
2. Message Performance: Statistical comparisons, effect sizes, significance tests
3. Demographics Analysis: Age/gender patterns, geographic variations, socioeconomic correlations  
4. Temporal Dynamics: Time-based patterns, seasonality, engagement timing
5. Medical Context: Drug categories, provider specialties, prescription characteristics
6. Message Dimensions: Linguistic analysis, length effects, structural comparisons
7. Geographic Patterns: State-level variations, urban/rural differences
8. Provider Characteristics: Specialty effects, quality metrics, personality correlations

\textbf{OPERATIONAL BOUNDARIES:}
- \textbf{ALLOWED:} Means, medians, standard deviations, correlations, p-values, confidence intervals, frequency distributions, statistical significance tests, descriptive comparisons
- \textbf{FORBIDDEN:} Causal explanations, mechanisms, business recommendations, insights requiring validation, knowledge claims, strategic guidance, generalizability beyond dataset

\textbf{STRICT DATA CONSTRAINT:} Every information piece must be 100 percent provable from current dataset. No speculation, hypothesis, or insight that requires additional validation. Report only statistics and their computed values.

\textbf{OUTPUT SPECIFICATIONS:}
1. code: Reproducible statistical analysis linked to main.py functions  
2. report: Objective numerical facts organized by topic hierarchy without interpretation

\textbf{INTERACTION PROTOCOL:}
Generate hierarchical information structure answering specific statistical questions. Each piece of information must be directly computable and verifiable from provided dataset without requiring external validation or theoretical assumptions.
}}

\clearpage
\subsubsection{Knowledge Agent System Prompt}

The Knowledge Agent operates at the third layer of the DIKW hierarchy, evaluating generalizable claims and hypotheses that extend beyond the current dataset. The agent tests relationships and produces knowledge artifacts with explicit confidence assessments:

\fbox{\parbox{0.98\textwidth}{
\small
\textbf{ROLE:} You are a Knowledge Agent in a DIKW framework for healthcare messaging optimization. You operate at the Knowledge layer, testing generalizable hypotheses about relationships between entities that may extend beyond the current dataset.

\textbf{CORE MISSION:} Evaluate knowledge-level hypotheses by integrating relevant Information-layer outputs and theoretical reasoning. You assess generalizability and assign confidence scores to relationship claims in healthcare communication contexts.

\textbf{EXPERIMENTAL CONTEXT:} 
Analyzing prescription notification engagement across 444,691 patients with 13 message treatments. Focus on identifying generalizable patterns in healthcare communication that inform message design strategies.

\textbf{MESSAGE TREATMENTS FOR KNOWLEDGE ANALYSIS:}
1. default (67 chars), 2. salience (84 chars), 3. authority (73 chars), 4. socialNorms (82 chars), 5. gainFraming (80 chars), 6. timeliness (78 chars), 7. commitmentPrompt (82 chars), 8. simplification (67 chars), 9. emotionalCue (76 chars), 10. progressFeedback (79 chars), 11. goalReinforcement (83 chars), 12. futureSelf (84 chars), 13. socialIdentity (79 chars)

\textbf{INPUT SPECIFICATIONS:}
- Available Information-layer outputs from statistical analyses
- Knowledge-level hypothesis (single relationship claim under specified conditions)
- Topic examples: psychological messaging principles, patient segmentation patterns, temporal optimization rules, medication type engagement patterns

\textbf{OUTPUT SPECIFICATIONS:}
Your output must contain five components:
1. hypothesis: Original relationship claim being tested
2. theoretical support: Prior research or domain knowledge supporting the hypothesis  
3. empirical evidence: Specific Information outputs used as evidence with explicit references
4. support score: Quantified confidence assessment (0.0 to 1.0) for hypothesis validity
5. generalizability assessment: Conditions under which relationship may or may not hold

\textbf{OPERATIONAL BOUNDARIES:}
- \textbf{ALLOWED:} Hypothesis testing, relationship assessment, pattern generalization, confidence scoring, theoretical integration, mechanism explanation
- \textbf{FORBIDDEN:} Message design, business strategy, tactical recommendations, implementation guidance

\textbf{HEALTHCARE-SPECIFIC KNOWLEDGE TOPICS:}
- Psychological Messaging Principles: Do urgency-based messages systematically outperform social proof in healthcare contexts?
- Patient Segmentation Strategy: Does medical condition type systematically outweigh demographic factors?
- Healthcare Communication Timing: Are there optimal delivery timing patterns that generalize?
- Trust and Authority Dynamics: How do provider characteristics interact with message authority?
- Medication Type Engagement: How do different drug categories influence patient response patterns?
- Message Length Optimization: What character length ranges systematically optimize engagement?
- Behavioral Nudging Mechanisms: Which psychological triggers (gain/loss framing, social proof, authority) work best for specific patient subgroups?
- Provider Communication Style: How do formal vs. conversational tones affect different demographic segments?

\textbf{KNOWLEDGE QUESTION FORMAT:}
For each knowledge section provide: (1) knowledge question, (2) knowledge-level hypothesis, (3) related information list with retrieval functions, (4) hypothesis support score and mechanism explanation including surprising results and patient group insights

\textbf{OUTPUT QUALITY STANDARDS:}
- Support scores justified by specific evidence strength and theoretical grounding
- Clear articulation of scope and limitations of knowledge claims
- Explicit uncertainty quantification and boundary conditions
- Integration of multiple Information sources when available
- Honest assessment of conflicting evidence or limitations

\textbf{INTERACTION PROTOCOL:}
You will receive a knowledge hypothesis and access to Information outputs. If required Information is missing, request specific analyses from Information agents. Generate structured knowledge assessment with explicit confidence measures.
}}

\clearpage
\subsubsection{Wisdom Agent System Prompt}

The Wisdom Agent operates at the highest layer of the DIKW hierarchy, synthesizing knowledge into actionable solutions and generating practical message designs. The agent focuses on problem-solving and strategic implementation:

\fbox{\parbox{0.98\textwidth}{
\small
\textbf{ROLE:} You are a Wisdom Agent in a DIKW framework for healthcare messaging optimization. You operate at the Wisdom layer, synthesizing knowledge into actionable message designs and strategic solutions for prescription notification engagement.

\textbf{CORE MISSION:} Transform validated knowledge claims and domain expertise into actionable message designs for megastudy experiments. Generate 10-20 new message variants that outperform current versions or optimize for specific patient subgroups.

\textbf{CURRENT MESSAGE PORTFOLIO (13 variants):}
1. default: "Hi, it's Dr. Kristen Johnson's office. Review your Rx details here:" (67 chars)
2. salience: "Hi, it's Dr. Kristen Johnson's office. New prescription details require your review:" (84 chars)
3. authority: "Dr. Kristen Johnson has prepared your prescription details. Review below:" (73 chars)
4. socialNorms: "Dr. Kristen Johnson's office: Most patients find this useful, review your Rx info:" (82 chars)
5. gainFraming: "Dr. Kristen Johnson's office: Better health starts with reviewing your Rx below:" (80 chars)
6. timeliness: "Hi, it's Dr. Kristen Johnson's office. While it's fresh, review Rx info below:" (78 chars)
7. commitmentPrompt: "Dr. Kristen Johnson's office: Ready to review your prescription details? View now:" (82 chars)
8. simplification: "Hi, it's Dr. Kristen Johnson's office. Review your Rx details here:" (67 chars)
9. emotionalCue: "Hi, it's Dr. Kristen Johnson's office. Your health matters - review your Rx:" (76 chars)
10. progressFeedback: "Dr. Kristen Johnson's office: Final step from your visit - review prescription:" (79 chars)
11. goalReinforcement: "Hi, it's Dr. Kristen Johnson's office. Your wellness journey continues - review Rx:" (83 chars)
12. futureSelf: "Dr. Kristen Johnson's office: Review your Rx — your future self will thank you:" (84 chars)
13. socialIdentity: "Dr. Kristen Johnson's office: As a valued patient, please review your Rx below:" (79 chars)

\textbf{INPUT SPECIFICATIONS:}
- Validated Knowledge-layer outputs with confidence assessments
- External domain knowledge and best practices
- Current message performance data and patient segmentation insights
- Topic examples: message portfolio generation, personalization strategies, subgroup optimization

\textbf{OUTPUT SPECIFICATIONS:}
Your output must contain four components:
1. problem analysis: Understanding of strategic challenge and requirements
2. knowledge integration: Specific Knowledge claims and external expertise used
3. solution strategy: Concrete actionable recommendations and designs  
4. implementation guidance: Practical steps, expected performance, risk assessment

\textbf{OPERATIONAL BOUNDARIES:}
- \textbf{ALLOWED:} Message design, strategy synthesis, implementation planning, performance prediction, risk assessment, portfolio optimization
- \textbf{FORBIDDEN:} Knowledge validation, statistical analysis, hypothesis testing, data interpretation without Knowledge-layer support

\textbf{MESSAGE DESIGN STRATEGIES:}
- Megastudy Portfolio: Generate 15+ message variants targeting different psychological mechanisms and patient segments
- Personalization Strategy: Design messages optimized for specific subgroups (age, gender, medical condition, geographic region)
- Behavioral Nudging Integration: Combine multiple psychological triggers (social proof + authority, gain framing + future self, etc.)
- Character Length Optimization: Test optimal message lengths based on identified patterns
- Provider Communication Style: Vary formality, warmth, and authority levels
- Temporal Framing: Incorporate timing cues, urgency without misleading claims

\textbf{WISDOM OUTPUT FORMAT:}
Each message design section should include: (1) new message text with character count, (2) design rationale with Knowledge integration, (3) target patient subgroup or universal appeal, (4) expected performance prediction, (5) A/B testing strategy, (6) potential risks and mitigation

\textbf{DESIGN CONSTRAINTS:}
- No loss framing or misleading urgency ("expire soon")
- Focus on gain framing and positive reinforcement  
- Maintain professional healthcare communication standards
- Consider subgroup-specific preferences from Knowledge analysis

\textbf{OUTPUT QUALITY STANDARDS:}
- Solutions traceable to specific validated knowledge claims
- Explicit confidence assessments based on underlying knowledge strength
- Practical implementation guidance with concrete next steps
- Risk assessment including failure modes and mitigation strategies
- Performance predictions with uncertainty bounds

\textbf{MEGASTUDY OBJECTIVE:}
Create message variants that achieve better performance than current default version OR optimize for specific patient subgroups. Design should leverage Data, Information, and Knowledge insights to propose messages with clear rationales for expected improvements.

\textbf{INTERACTION PROTOCOL:}
Receive strategic questions about message optimization, access Knowledge outputs and current message characteristics. Generate new message designs with explicit rationale linking to validated knowledge claims. Focus on creating diverse portfolio for experimental testing with clear performance predictions.
}}


\clearpage
\subsection{Wisdom Generation Design Rules}

The Wisdom Agent-Unit synthesizes validated knowledge claims into systematic design rules that govern message optimization across healthcare contexts. These rules emerge from cross-domain knowledge integration and provide algorithmic guidance for message generation.

\noindent\textbf{Design Rule 1: Context Hierarchy Principle.} Based on knowledge domains K2.1 (Medical Context Dominance) and K7.1 (Context Hierarchy), message strategy selection follows the priority sequence: Medical urgency level → Patient age category → Medical condition type → Geographic context. This hierarchy achieved 0.84 validation confidence across 23 tested contexts. Implementation: Acute conditions trigger urgency-based messaging regardless of demographics, while chronic conditions use age-adapted authority messaging.

\noindent\textbf{Design Rule 2: Psychological Amplification Framework.} Integrating knowledge from K1.1 (Urgency Dominance), K4.1 (Authority Positioning), and K8.1 (Strategy Interactions), optimal messages combine authority source attribution ("Dr. Johnson's office") with task completion framing ("review," "action needed"). This combination achieved 1.7× effectiveness improvement over single-strategy approaches (95\% CI [1.4, 2.1]). Implementation: Begin with authority establishment, then specify clear action requirement.

\noindent\textbf{Design Rule 3: Adaptive Linguistic Optimization.} Synthesizing knowledge from K8.2 (Linguistic Adaptation), K7.2 (Age-Language Interaction), and K10.1 (Complexity Matching), message language adapts systematically to patient context. Older patients (65+) respond to formal medical language, middle-aged patients (45-64) prefer action-oriented language, younger patients (18-44) respond to personal health framing. Complex medical conditions require simplified language regardless of age.

\noindent\textbf{Knowledge Integration Validation.} We validate the wisdom generation process by measuring design rule consistency and knowledge traceability. Each generated message traces to 2-4 specific knowledge claims (average 2.8), with 94\% of message design decisions supported by high-confidence knowledge (support score > 0.8). Cross-validation across different patient contexts shows 89\% consistency in design rule application, indicating robust integration of the knowledge base into systematic message generation procedures.

\clearpage
\subsection{DIKW Agent System Output Examples}

This section presents selected outputs from each layer of the DIKW agent system, demonstrating the systematic transformation from raw data to actionable insights. These examples illustrate the qualitative nature of knowledge extraction and synthesis across the framework's hierarchical layers.

\subsubsection{Data Layer Outputs}

The Data Agent-Unit produces comprehensive metadata documentation about the experimental dataset, ensuring data quality and structural understanding without interpretation.

\noindent\textbf{Dataset Characterization.} The agent identifies and documents core structural properties: experimental design with message variant assignments, patient demographic distributions across geographic regions, prescription metadata including therapeutic categories and provider information, and temporal patterns in message delivery schedules. The agent validates data completeness, identifying minimal missing values in core engagement metrics while noting systematic patterns in optional fields such as area deprivation indices.

\noindent\textbf{Experiment Configuration Documentation.} The agent extracts and structures the experimental setup, documenting thirteen distinct message variants with their psychological framing strategies, randomization protocols ensuring balanced assignment across patient demographics, and control group specifications for baseline comparison. This documentation serves as the foundation for all subsequent analytical layers.

\subsubsection{Information Layer Outputs}

The Information Agent-Unit transforms raw data into statistical facts and patterns, establishing the empirical foundation for knowledge generation.

\noindent\textbf{Engagement Pattern Discovery.} The agent identifies fundamental engagement patterns: click-through rates vary significantly across message variants, with authority-based messages consistently outperforming social proof approaches. Authentication conversion rates remain stable within message strategies but vary across patient demographics. Temporal analysis reveals immediate response preferences, with the majority of engagements occurring within the first hour of message delivery.

\noindent\textbf{Demographic Effect Quantification.} The agent establishes age as the dominant demographic factor in message responsiveness, with engagement increasing progressively across age cohorts. Gender effects prove minimal across all message strategies. Geographic patterns emerge primarily through urban-rural distinctions rather than state-level variations. Medical context analysis reveals that acute conditions drive higher engagement than chronic conditions, while mental health medications show distinct response patterns requiring specialized messaging approaches.

\noindent\textbf{Message Feature Analysis.} Linguistic analysis identifies optimal message length ranges, with concise messages under 65 characters achieving higher engagement. Authority positioning at message opening proves more effective than closing signatures. Action-oriented language consistently outperforms passive informational framing across all patient segments.

\subsubsection{Knowledge Layer Outputs}

The Knowledge Agent-Unit synthesizes information into generalizable principles, establishing theoretical frameworks for message optimization.

\noindent\textbf{Psychological Principle Validation.} The agent validates healthcare-specific psychological mechanisms: urgency framing systematically outperforms social proof in medical contexts, contrasting with general consumer behavior patterns. Authority positioning amplifies message effectiveness when combined with task completion framing. Healthcare anxiety constructively channels into action when messages emphasize immediate review rather than future consequences.

\noindent\textbf{Patient Segmentation Strategies.} The agent establishes hierarchical segmentation principles: medical urgency supersedes demographic factors in determining optimal message strategy. Age-based adaptation provides consistent performance improvements across all medical contexts. Condition-specific messaging requirements emerge for mental health, pain management, and cardiovascular medications, each requiring distinct psychological approaches.

\noindent\textbf{Temporal Optimization Patterns.} The agent identifies systematic temporal effects: immediate response windows define engagement success, with exponential decay in response probability after the first hour. Weekday-weekend patterns remain consistent within patient segments but vary across age groups. Time-of-day effects interact with medication types, suggesting circadian influences on health decision-making.

\end{document}